\documentclass[conference]{IEEEtran}
\IEEEoverridecommandlockouts

\usepackage{etoolbox}
\newbool{draft} \setbool{draft}{false}

\usepackage[english]{babel}
\usepackage{graphicx}
\usepackage{hyperref}
\usepackage{amsfonts}
\usepackage{amsmath}
\usepackage{amssymb}
\usepackage{todonotes}
\usepackage{xcolor}
\usepackage{algorithmicx}
\usepackage{algpseudocode}
\usepackage{multirow}
\usepackage{caption}
\usepackage[backend=biber,style=ieee]{biblatex}
\title{Zero-Shot Hierarchical Classification on the Common Procurement Vocabulary Taxonomy}

\ifbool{draft}{}{\author{\IEEEauthorblockN{Federico Moiraghi}
\IEEEauthorblockA{\textit{University of Milano Bicocca} \\
\textit{SpazioDati s.r.l.} \\
Italy \\
\href{mailto:f.moiraghimotta@campus.unimib.it}{f.moiraghimotta@campus.unimib.it}}
\and

\IEEEauthorblockN{Matteo Palmonari}
\IEEEauthorblockA{\textit{University of Milano Bicocca} \\
Italy \\
\href{mailto:matteo.palmonari@unimib.it}{matteo.palmonari@unimib.it}}
\and

\IEEEauthorblockN{Davide Allavena}
\IEEEauthorblockA{\textit{SpazioDati s.r.l.} \\
Italy \\
}
\and

\IEEEauthorblockN{Federico Morando}
\IEEEauthorblockA{\textit{SpazioDati s.r.l.} \\
Italy \\
}

}
}

\begin{document}

\maketitle

\begin{abstract}\footnote{Full-length version of the short paper accepted at \href{https://ieeecompsac.computer.org/2024/}{COMPSAC 2024}.}
Classifying public tenders is a useful task for both companies that are invited to participate and for inspecting fraudulent activities. To facilitate the task for both participants and public administrations, the European Union presented a common taxonomy (\textit{Common Procurement Vocabulary}, CPV) which is mandatory for tenders of certain importance; however, the contracts in which a CPV label is mandatory are the minority compared to all the Public Administrations activities.
Classifying over a real-world taxonomy introduces some difficulties that can not be ignored. First of all, some fine-grained classes have an insufficient (if any) number of observations in the training set, while other classes are far more frequent (even thousands of times) than the average. To overcome those difficulties, we present a zero-shot approach, based on a pre-trained language model that relies only on label description and respects the label taxonomy.
To train our proposed model, we used industrial data, which comes from \url{contrattipubblici.org}, a service by \href{https://spaziodati.eu}{SpazioDati s.r.l}. that collects public contracts stipulated in Italy in the last 25 years. Results show that the proposed model achieves better performance in classifying low-frequent classes compared to three different baselines, and is also able to predict never-seen classes.
\end{abstract}

\section{Introduction}
``Accounting for a trading volume of EUR 2.448 billion (approx. 16\% of the 2017 EU GDP), European
public procurement is a major driver for economic growth, job creation, and innovation''\footnote{Quoted from a study titled ``Contribution to Growth European Public Procurement'' published on the official European Parliament website and available online at \url{https://www.europarl.europa.eu/RegData/etudes/STUD/2018/631048/IPOL_STU(2018)631048_EN.pdf}.}.  
Since public tenders are advertised on different European or national institutional websites,  many companies have developed solutions to monetize simplified and improved access to knowledge about public tenders at scale.
These solutions are usually based on processing crawled data to improve their quality and make them accessible in different modalities (query, recommendation, subscriptions, dashboards, etc.); typical processing operations include cleaning, interlinking, enriching with additional metadata, storing, and so on.
A crucial task to improve access is to classify the data properly. 

The \textit{Common Procurement Vocabulary} (CPV) has been developed by the European Union to harmonize the classification of public tenders, which covers multiple domains and human activities, and 
\textit{``facilitate the processing of invitation to tender published in the Official Journal of the European Union (OJEU)''}\footnote{Definition from \href{https://en.wikipedia.org/wiki/Common_Procurement_Vocabulary}{wikipedia}.}. 
The last version (2008) of the taxonomy has $9454$ labels, in a tree that has a maximum depth of seven\footnote{The taxonomy and more information are available on the official website: \url{https://simap.ted.europa.eu/web/simap/cpv}.}.
Each CPV label is represented by a code of 8 digits and a control code (for manual verification);
a natural language description is also attached to every label (in 24 different languages) to simplify interpretability by humans.
For many tender data published online, the CPV labels are not available, not accurate, or too generic. Improving the capability of classifying public tenders based on CPV taxonomy is relevant to improving access to these data in both the public and industry sectors.
In this paper, we discuss an approach aimed at classifying unlabeled public tenders to improve their recommendation to customers with more precise information.

\begin{figure}[t]
    \centering
    \includegraphics[width=1.2\linewidth]{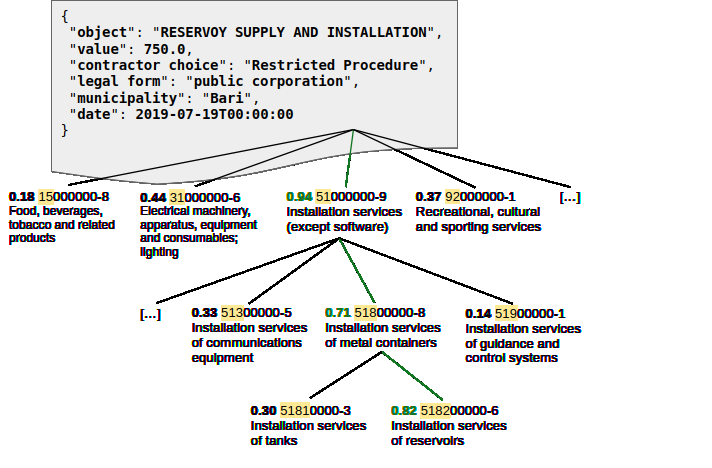}
    \caption{An example of an input document (above), composed by the main field \texttt{"object"} and other meta-data, and a portion of the taxonomy used for classification (below). Each label (numerical code) has a canonical description in 24 different languages. Notice \textit{``reservoy''} instead of \textit{``reservoir''}: such errors are quite common.}
    \label{fig:example}
\end{figure}

The classification of tender with the CPV taxonomy is a challenging problem for several reasons. First, data available about tenders are limited: a sample of this data is provided in the box in Figure~\ref{fig:example}. 
Second, classifying over a taxonomy is more challenging than classifying with ``flat'' labels: the target label may be a not-leaf node, enormously increasing the output size. There are also some natural correlations between labels that can not be ignored (it may be suitable to classify an observation as a parent or descendant of the true label). 
In addition, in the CPV taxonomy, many fine-grained labels are seldom or never used, thus making it difficult to find training examples for a large number of labels (e.g. ``sesame oil'' is hardly ever used, since foodstuff is usually acquired in a more general way and not by single products).
Classifying over a taxonomy, considering also non-leaf nodes, is a sub-task referred to as \textit{not leaf-mandatory hierarchical classification}~\cite{silla2011survey}.
This task has two main difficulties: iterative mechanisms may be needed to classify observations into classes of increased complexity, as sketched in Figure~\ref{fig:example}, and to understand when it does not make sense to keep on exploring finer-grained labels;
it should have a system to avoid error propagation (an error in a certain point of the tree implies an error in all the following predictions).

This work investigates the exploitation of a
bi-directional Pre-trained Language Model (PLM)~\cite{devlin2019bert} to improve the classification of tender data into CPV taxonomies taking inspiration from the process that a human makes: the task is easier with a direct comparison of the tender object with the label description.
Our main contribution consists of proposing an architecture for zero-shot entity linking \cite{wu2019scalable} to support zero-shot classification iteratively over the taxonomy.


A practical use case for a model that can predict the CPV of a public tender is filling missing-value of open knowledge bases about invitations to tender such as \href{https://theybuyforyou.eu}{TheyBuyForYou} \cite{soylu2022theybuyforyou} or \href{https://contrattipubblici.org}{ContrattiPubblici}\footnote{Made by SpazioDati: \url{https://spaziodati.eu}} \cite{futia2018contrattipubblici}.
Those data are useful in fraud detection (since in some circumstances some information is not mandatory for the records) as well as improving integration between European Countries \cite{info13020099}.

\section{Related Works}\label{sec:related}
Since CPV prediction is a task not well explored in literature, we divided this section in two: the first part shows some references that are useful in understanding both the task and the model development, presenting the difficulties and strategies of hierarchical (zero-shot) classification.
The second part instead shows previous works on the CPV prediction task, showing why a zero-shot system may be of interest.

\subsection{Hierarchical Classification}
Classifying through a taxonomy is a sub-task of HC, a task well analyzed by \citeauthor{silla2011survey} \cite{silla2011survey}.
They define the task as a supervised classification through a taxonomy that is defined \textit{a priori} (and not generated on-the-fly).
The difference with \textit{Structured Classification} is the fact that HC defines the structure between labels specifically as a tree or a DAG (\textit{Direct Acyclic Graph}) and not in free form.
Notice that HC output can be either multi-label or single-label classification.
HC is not only aimed at making predictions when a formal taxonomy is explicitly defined in the dataset (e.g. \cite{zhong2023attention} which uses an attention-based model for music genre prediction according to a commercial taxonomy, \cite{PARMEZAN2022101751} that classify insects according to flying noise, dividing between pollinators and not pollinators) but also when the taxonomy exists and is (in general) accepted by domain experts: for example, \cite{kahl2019overview} propose a dataset for bird recognition without providing the taxonomy, however biological taxonomy (although discussed and modified by biologists) can be accepted.

Depending on the dataset structure and aim, HC models may or may not predict leaf labels; in addition, various strategies to exploit the tree structure have been proposed in literature.
The most common is the \textit{top-down} approach, which consists of a series of classifiers each one predicting the next step (with growing granularity); on the opposite direction, \textit{bottom-up} (not well explored in literature) and \textit{big-bang} have a classifier which predicts respectively the leaf labels (whose probabilities are then combined) or the taxonomy as a whole.
Both have their pros and cons: \textit{bing-bang} models reduce error propagation given by error multiplication, while on the other hand \textit{top-down} models are more computationally expensive yet have fewer labels to manage (per sub-model).
There are also other new or hybrid approaches: for example, \cite{zhou-etal-2020-hierarchy} proposes a tree architecture that hard-codes the taxonomy; this approach however considers only leaf labels.
\cite{pmlr-v80-wehrmann18a} uses a custom loss to penalize taxonomy violations in a flat classifier for protein-function prediction (a multi-label task).
\cite{Bertinetto_2020_CVPR} does the same thing on a Computer Vision multi-label classification task in order to obtain a better list of tags.
As far as we know, those model-based approaches have not been tested yet on a document classification dataset.
\newline

This task has some peculiar difficulties compared to flat classification.
\paragraph{Error Propagation.}
In a top-down model, if a sub-classifier makes a wrong classification at some point, all the subsequent classifications are necessarily wrong, and therefore the error is propagated through the taxonomy.
A simple solution may be using a single \textit{big-bang} model which considers all the output labels (either leaves only or the whole taxonomy); however, this approach is possible only with small taxonomies.
To mitigate the error, \cite{hernandez2014multidimensional} proposes a model that considers multiple root labels first and then expands them all for a single-label classification task; this strategy however does not improve the results.
\cite{ramirez2016hierarchical} instead uses a system for providing each solution sub-tree a probability score.
\cite{guo2021hierarchical} expands the idea, considering at each step the first $k$ results for the expansion.
We opted for a similar approach, considering all the classes with a score above a given threshold.
With a completely different approach, \cite{liu2022hierarchical} adds to each sub-model a fake class that represents the label parent: if this class is predicted, the second (or third, fourth...) best result of the previous prediction is explored instead; this is possible since their model stops on a leaf label.

\paragraph{Stopping Strategy.}
This feature of an HC system is crucial when non-leaf labels are allowed: it checks if the classification should go deeper or not, returning the last results as output.
A good stopping may, in addition, mitigate error propagation since it may claim that the classification is wrong. \newline
While a common strategy is a threshold method (e.g. \cite{parmezan2020combination}), there are also other more sophisticated approaches.
\cite{ramirez2016hierarchical} uses a threshold over a combined probability in order to decide when to stop the classification.
\cite{a10040138} uses a simpler approach: it stops when the probability of a child's label is lower than his parent's; the idea is that as granularity increases, the model should be more certain about the result.
With a more sophisticated approach, \cite{WANG2022644} presents a binary classifier that considers label probability distribution to decide if going deeper or not.
We took inspiration from this last work, using a small neural network instead of a logistic regression.

\paragraph{Sampling Strategy.}
Finally, a particularity of hierarchical classification is that, even in single-label classification, an observation may belong to multiple classes with parent-child relationships (e.g. \textit{all cats are mammals}).
This task characteristic imposes the choice of a \textit{sampling strategy}, which shows positive and negative examples for each classifier.
\cite{eisner2005improving, fagni2007selection, ceci2007classifying} show six different strategies.  
However, \cite{8575650}, after extensive experimentation, demonstrates that there is no best strategy for sampling, but it must be chosen according to task semantics.
After some preliminary tests, we opted to train our model using the \textit{exclusive sibling} strategy, since it is conceptually similar to \textit{hard negative training} used to train BLINK \cite{wu2019scalable}, which uses a model similar to ours.
\newline

For zero-shot hierarchical document classification, \cite{chalkidis2020empirical} uses a pre-trained BERT encoder in order to obtain a latent representation of the label; then another BERT instance actually makes the classification aligning the latent representation of the input with the  associated labels, in a bi-encoder fashion.
Compared to us, it needs long textual descriptions of the labels to obtain a good latent representation, while in our taxonomy most labels have very short descriptions, even one word long.
\cite{yiExploringHierarchicalGraph2022} instead suggests that simpler pre-trained embeddings may not be sufficient for fine-grained classification (on an image-classification task): they propose a method for obtaining better embeddings fine-tuning a transformer-based model.
This method, however, uses a \textit{big-bang} strategy for a multi-label task, with poor performances on the first candidate (with a strong improvement considering also the followings) and therefore we think it is not well suited for single-label classification.
Compared to our approach, those methods define a fixed representation of the label in a common space, being more interested in multi-label than single-label classification.

\subsection{CPV Prediction}
Classification through the \textit{Common Procurement Vocabulary} is a task little explored  in literature.
The most recent and more complete work about this niche domain is \cite{navas-loroMultilabelTextClassification2022a}, which focuses on the Spanish language and approaches the task as both structured and \textit{Extrime Multi-Label Text Classification} (XMTC) task; however, their model is limited to root predictions (45 classes).
Their approach is based on a RoBERTa \cite{liuRoBERTaRobustlyOptimized2019} classifier; according to the authors, they suffered from data unbalance in the training data.
Also \cite{kayte2019mixed, ahmia2020assisted} predict only the root labels using SVM models, while \cite{MKaan} uses a BERT model.
As far as we know, there is no work that considers the whole CPV taxonomy.
The only work that considers also (a portion of) non-roots classes is \cite{deloitte}, a technical business report that claims the existence of a model that can predict the first four digits (roots and two levels above) \textit{``with an
accuracy of 70\%''} for a new product; however, according to \cite{navas-loroMultilabelTextClassification2022a}, they manually engineer the taxonomy.

\section{CPV Taxonomy \& the Dataset}
The CPV taxonomy is composed of 9454 classes (6531 of which are leaves), belonging to 45 main labels.
The average number of children per class is $1.00 \pm 1.99$, until a maximum depth of $7$.
In the whole database, which has $6.7e7$ contracts recorded, 492 CPV labels have never been observed, and more than half (5366 labels, affecting about $4.5e6$ observations) have less than 100 records. \newline
\begin{figure}[h!]
    \centering
    \includegraphics[width=0.49\linewidth]{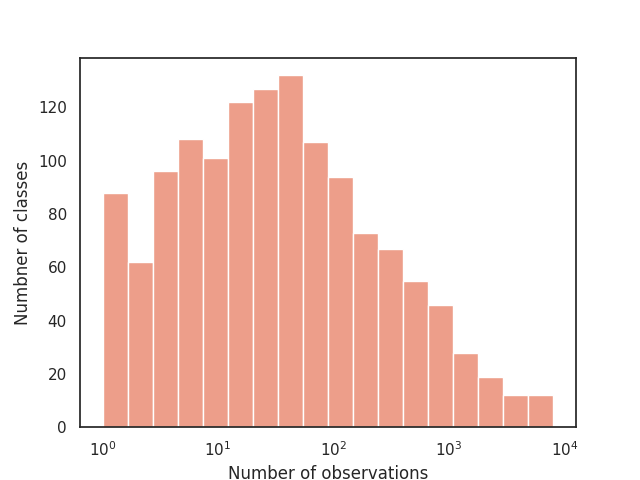}
    \includegraphics[width=0.49\linewidth]{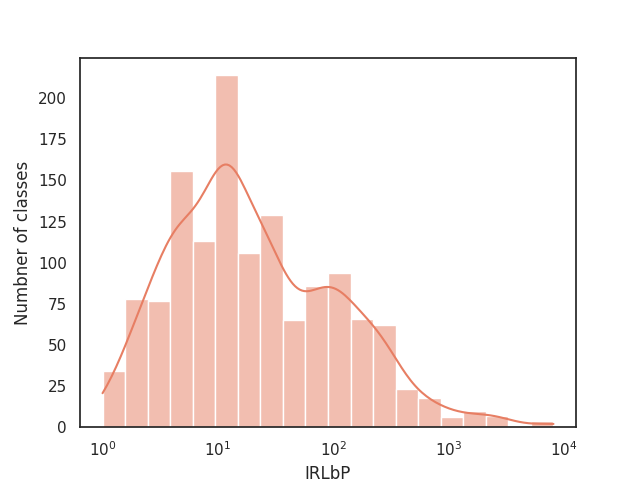}
    \includegraphics[width=0.49\linewidth]{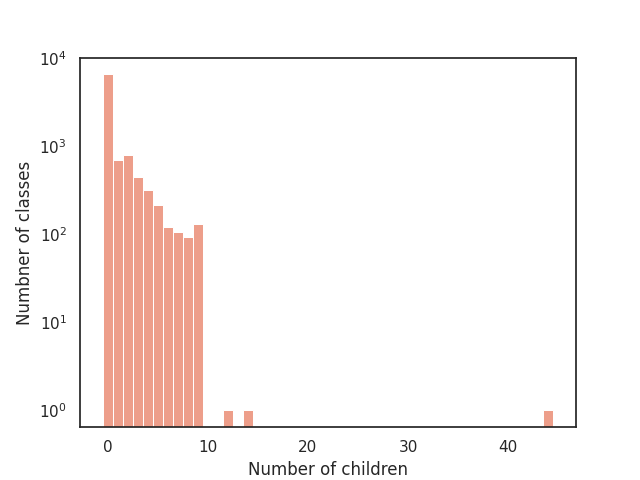}
    \includegraphics[width=0.49\linewidth]{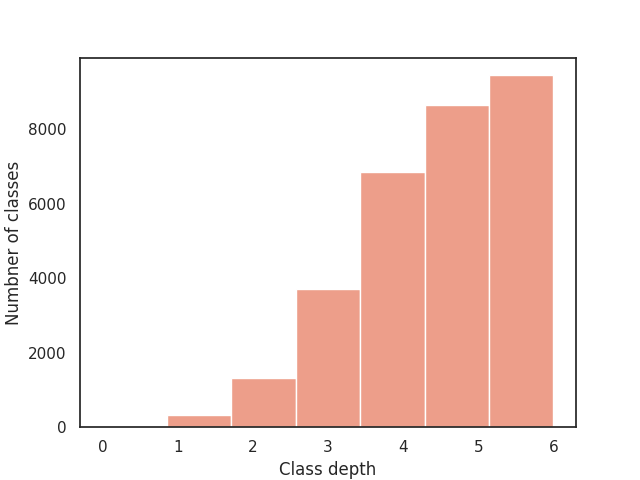}
    \includegraphics[width=0.70\linewidth]{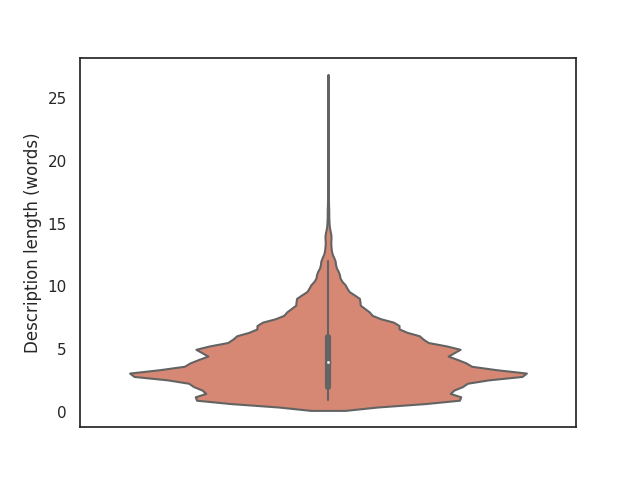}
    \caption{Some descriptive metrics about both the taxonomy and the dataset:
    frequency of the labels in the training data (top left, logarithmic scale);
    IRlBP distribution (top right, logaritmic scale);
    number of children per node (middle left, logarithmic scale);
    cumulative number of classes per depth (middle right, cumulative);
    number of words per label description (bottom).}
    \label{fig:taxonomy}
\end{figure}

Our task is to predict the CPV code for a contract, given its object and some meta-data (such as month, value, and typology of contract parties).
All the meta-data are categorical except for the value field, which is numeric and positive by definition: for this reason, this value was pre-processed to make it readable by a language model, replacing it with a special token representing the magnitude order (\verb|€| repeated as many time as the digits of the integer part of the value, with a maximum of 9).
For example, a contract with a value of €$1000.00$ is represented by the special token \verb|[€€€€]| (which is the same representation for €$9000.00$).

The main difficulties of the task are the fact that classes are very imbalanced and that not all the labels are included in the training data.
The fact that an observation may also be labeled with all the ancestor classes implies that it is not trivial to calculate how unbalanced classes may be.
For this reason, \cite{pereiraHierarchicalClassificationImbalanced2021} proposes a metric to calculate the imbalance for each class and for the whole dataset\footnote{We use the same mathematical notation of\cite{silla2011survey}.}:
\begin{equation*}
\begin{aligned}
 \text{IRLbP}(Y) &= \frac{\underset{Y^\prime \in \Upsilon}{\max} \sum^{|D|}_{i=1} h(Y^\prime, Y_i)}{\sum^{|D|}_{i=1} h(Y, Y_i)} \hspace{10pt} \in [1, +\infty) \\
    \text{HMeanIR} &= \frac{1}{|Y|} \sum_{Y_i}^{\Upsilon} \text{IRLbP}(Y_i) \hspace{10pt} \in [+1, +\infty)
\end{aligned}
\end{equation*}
Where $D$ is  the dataset, $Y$ is the set of labels (and $\Upsilon$ the taxonomy), $h(Y, Y_i) = 1 \Leftrightarrow Y \preceq Y_i$.
IRLbP score is essentially the ratio between the frequency of the most-represented class $Y^\prime$ and of the target class $Y$ in the dataset $D$: $\text{IRLbP}(Y) = 1$ means that the class is the most frequent one, higher the score higher the imbalance.
In our dataset, HMeanIR is equal to $119.16$ (meaning that the most frequent class is $119.16$ times more populated than the average frequency of the classes); IRLbP distribution can be seen in Figure \ref{fig:taxonomy}.

In addition, annotation granularity is not always as specific as the declared labels in the dataset: a contract that only says \textit{``foodstuff''} may have a label that indicates a precise kind of food (like, for example, ``\texttt{03210000-6} potatoes''); on the other hand a contract that indicates very specific medical equipment (such as a specific medicine or an ECG machine) may is labeled just as ``\verb|33000000-0| Medical Equipment''.
Other contract objects are not understandable even by a human but should be inferred by a context that is not included in the dataset (e.g. previous contracts of the same PA or additional information provided by the notice board in the physical office); an example may be \textit{``Results of the public notice May 2018.''} or \textit{``Payment invoice 23/0645''}.

This kind of noise can not be labeled as \textit{label noise} \cite{frenay2013classification} since, for this task, hyper-specific labels, as well as generic labels, are considered true; while insufficient information is not a kind of noise at all.
However, those dataset characteristics reduce drastically the model performances either because incomplete information is included in the dataset or because the evaluation is not well suited.
However, for the current work, we are more interested in comparing the proposed zero-shot approach with a traditional hierarchical classifier, and therefore raw performances are not a concern.

\section{Hierarchical Classification with Cross-encoder}\label{sec:model}
\nocite{wolf-etal-2020-transformers}
To achieve zero-shot by design we propose a Hierarchical Cross-Encoder (HCE), which is inspired by cross-encoders used in entity linking and matching tasks~\cite{wu2019scalable}. A cross-encoder evaluates the similarity of an input pair $\langle x, y \rangle$ using a pre-trained language model to validate the couple as a sentence classification task.
Cross-encoders use bidirectional language models such, as BERT, which we use in our model.
Using a cross-encoder to support classification is inspired by the observation that both the main input features (the object of the contract) and the label description are written in a natural language.
Other meta-data are either left as category descriptions (if categorical) or quantized and introduced as special tokens (if numerical).
In this way, the task is presented to the model as a binary task, which \cite{rivolli2020empirical} presents as an effective strategy for hierarchical classification; this asset reduces also the impact of data imbalance: positive and negative labels are arbitrarily provided by us.
The idea of comparing the input with the output for hierarchical text classification is not new: \cite{wang2021fhtc} trains a model that indicates if a vector representation of the output fits the input. 
This approach, however, is presented as a \textit{few-shot} architecture (it requires examples for the vector encoder) and trains more than one model for the classification; we instead train only one model on the whole architecture, providing each time a different label description to it.

However, we intend to use a cross-encoder in a way that is different from the one proposed by \cite{wu2019scalable} for \textit{Named Entity Linking}: we are interested in evaluating the similarity between an instance of tender data and a label in the CPV taxonomy $\Upsilon$, and not between two entity descriptions.
In fact, our hierarchical approach differs in the candidates processed by the model: usually, a cross-encoder architecture is used to provide an accurate score for a set of candidates retrieved by a similarity function in a dense space.
In our case, this is not possible for three main reasons: the first is that our taxonomy has short label descriptions (even one word long) that are difficult to represent in a dense space; the second one is that is not trivial to include information about taxonomy structure in a latent representation.
Finally, some descriptions, especially generic ones, have negations (e.g. \textit{``Installation services (except software)''}) which complicates the building of a dense representation.
For the second problem, dividing the space into multiple local sub-problems (e.g. a dense space for each label, to represent its children) may be a solution; however, the benefits are considerable only if a label has a higher number of children (notice that the average number of children per node is low).
In our case, it is more efficient in both resources and inference time to analyze all the descending labels instead of retrieving a smaller sample.

\subsection{Encoding and Interpretation}
The input of the LM is therefore so composed: at the first position, after the \verb|[CLS]| special token, there is the contract object (or a reduced version to not overflow the 512 sub-word limit imposed by BERT architecture\footnote{This happens very hardly, about 1:10'000 times; and a human reader can easily understand the meaning from the first words. Therefore we think that an architecture specifically designed for long documents is not needed.}).
Then, each feature is preceded by a special token that represents its meaning (e.g. \verb|[MONTH] December|).
Notice that this architecture permits good missing-value management as well: if a feature is missing, it is just skipped.
Finally, a \verb|[SEP]| token precedes the label description.
Here is an example:
\begin{center}
{\footnotesize \verb|[CLS]| Food stuff \verb|[MONTH]| April \verb|[VALUE]| \verb|[€€€]| [\dots] \verb|[SEP]| Agricultural and horticultural products}
\end{center}
We preferred this setting over using only the \verb|[SEP]| special token since the validation of the couple improves a little (about $0.02 - 0.04$) in both precision and recall.
Notice that deep fine-tuning of the network has to be made in any case because we introduced special tokens for the representation of the \verb|value| field (which domain experts suggest is useful for the task).

The output is a real number in the interval $[0; +1]$ which represents the likelihood that the label description fits the input features; notice that, despite we worked on an Italian dataset, the approach works in the same way in any languages as well, as there are label descriptions.
As with any other cross-encoder, the output provides information only about one label and therefore the model requires more than one iteration per observation to predict between multiple labels, even if siblings: this is the main drawback of the proposed model.

\subsection{Training}
This input is then used to train the LM for a document classification task, to validate the couple $\langle x, y \rangle$.
In the training step, each observation has an equal probability of being presented as a positive or negative example, as well as an equal probability of being an instance of $Y_i \preceq Y_{\text{observed}}$. 
For the example in Figure \ref{fig:example}, the generation probabilities are summarised in Table \ref{tab:example_prob}.
\begin{table}
    \centering
    \begin{tabular}{|c|c|l|}
    \hline
        \multicolumn{2}{|c|}{Probability} & Generation Result \\
    \hline
    \multirow{3}{*}{$\frac{1}{2}$} & $^1/_3$ & a false root label (44 labels) \\
                          & $^1/_3$ & a false children of \verb|15000000-9| (9 labels) \\
                          &$^1/_3$ & $\langle$ \verb|51810000-3|, $\bot \rangle$ \\
    \hline
    \multirow{3}{*}{$\frac{1}{2}$} & $^1/_3$ &$ \langle$ \verb|51000000-9|, $\top \rangle$ \\
                          & $^1/_3$ & $\langle$ \verb|51800000-8|, $\top \rangle$ \\
                          & $^1/_3$ & $\langle$ \verb|51820000-6|, $\top \rangle$ \\
    \hline
    \end{tabular}
    \caption{Probabilities of the training generation process. $P(y = \top) = P(y = \bot)$, so that the output label is now balanced during the training process. The probability of generating a specific false label depends on the number of siblings: it becomes, for example: $P(\text{\texttt{15000000-8}}) = P(y=\bot) \cdot P(\text{depth}=0) \cdot \frac{1}{44} = 0.38\%$}
    \label{tab:example_prob}
\end{table}

Negative examples are generated with the \textit{exclusive sibling} strategy: randomly choosing one label between the node siblings.
This generation procedure is not fixed and is made at the beginning of each training epoch.
We trained the LM for five epochs using the default parameters provided by \verb|transformers| Python library by \href{https://huggingface.co/}{Huggingface}.
At the end of the training, the BERT model achieves an accuracy of $0.9441$ on the test set in recognizing if the couple is correct or not.

\subsection{Hierarchical Inference}
Our model is aimed at computing an advanced similarity function between the input and output by exploring the taxonomy: the inference procedure resembles the one used for a top-down local-per-node approach.
However, we do not exclude any path, as \cite{guo2021hierarchical,hernandez2014multidimensional} do, but we consider all labels whose probability reach a predefined threshold (set to $0.5$ in this work). 

We explore one taxonomy node at a time and compute confidence scores over its descendants. 
The exploration is made recursively until either a leaf node or some exit conditions are reached: either the label score is above a given threshold or a stopper algorithm, which considers score distribution, triggers the stop.
Notice that, if no root label reaches the given threshold, the HCE abstains from classification, since no label guarantees the desired confidence.
The inference process is described in detail in Algorithm 1: at the end of the process, it provides a list of labels descendingly sorted by score: we intend to retrieve multiple labels ordered for probability and not only the most probable one.

\begin{table}[h!]
\begin{algorithmic}
\Function{Predict}{$x$, $\Upsilon$, $label = \varnothing$, $score = \varnothing$}
    \If{$score < threshold$}
        \State{\Return{$\varnothing$}}
    \ElsIf{$\Call{leaf?}{label, \Upsilon}$}
        \State{\Return{$\{\langle label, score \rangle\}$}}
    \EndIf
    \If{$label = \varnothing$}
        \State{$candidates \gets \Call{roots}{\Upsilon}$}
    \Else
        \State{$candidates \gets \Call{children}{label, \Upsilon}$}
    \EndIf
    \State{$\hat{y} \gets \Call{cross-encoder}{x, candidates}$}
    \If{$label \ne \varnothing \wedge \Call{stopper}{\hat{y}}$}
        \State{\Return{$\{\langle label, score \rangle\}$}}
    \EndIf
    \State{$output \gets \bigcup_{c \in candidates} \Call{Predict}{x, \Upsilon, c, \hat{y}_c}$}
    \State{\Return{$\Call{sort by score}{output}$}}
\EndFunction
\end{algorithmic}
\caption*{Algorithm 1: Pseudo-code of the zero-shot classification pipeline.}
\label{alg:inference}
\end{table}

The reason to use a top-down pipeline is that the number of iterations is lower compared to a bottom-up or big-bang approach: the main drawback of an LM-based model, in this context, is its long inference time.
An advantage of our approach, on the other hand, is that we train and use a single model, which can be used with all the taxonomy labels.

\section{Experiments}\label{sec:experiments}
We are interested in analyzing if the proposed model reaches good performances in zero-shot classification, and on uncommon labels.
In fact, since labels are highly imbalanced, we are not interested in good micro-performances but in good performances on low-frequent or new classes.
As discussed in Section~\ref{sec:related}, previous methods for CPV prediction did not tackle the classification of tender data at any depth of the CPV taxonomy, but focused on the most generic classes (root nodes)~\cite{navas-loroMultilabelTextClassification2022a,kayte2019mixed, ahmia2020assisted,MKaan}.
As a consequence, they do not propose any solution to address CPV prediction as a not leaf-mandatory hierarchical classification problem, which is the main focus of our paper.
In addition, as \citeauthor{navas-loroMultilabelTextClassification2022a} stress, data are highly imbalanced and some labels are missing: we are interested in zero-shot classification.

\subsection{Data pre-processing}
In our experiments, we use some data transformations on the input that were found useful during exploratory analyses (omitted in this paper) to aggregate data and reduce input dimension.
In particular, we grouped the date in months and the municipalities in four macro-areas (North, Middle, South Italy \& Islands).
Despite being dataset-specific, municipalities aggregation can be easily generalized and adapted to other Countries; the HCE however is not affected by those transformations, receiving the data as is.
The value field instead is processed with a logarithmic transformation (since it is positive by definition, values equal to zero are considered missing).

\subsection{Baselines}
To make a valid comparison, we test our HCE with three baselines, also a contribution of ours, which makes the classification in three different ways using the same data as input.
In addition, we try to integrate the HCE with the most promising baseline for predicting unseen classes: this test aims to investigate if this model may be used with an existing one.

We introduce three \textit{baselines} which obtain a dense representation of the tender data and meta-data without relying on an LM (as made in \url{contrattipubblici.org} before transformer-based models become popular).
Then, we tested three main strategies to handle not leaf-mandatory hierarchical classification problems, which we discussed in Section~\ref{sec:related} and apply different approaches to explore the taxonomy. 

To extract \textit{initial features} of the input data, each baseline model considers a vector that is composed of (1)
\begin{itemize}  
   \item the tf-idf for each term in the object field (e.g., \textit{``Reservoy supply and installation''} in Figure~\ref{fig:example}) that appears at least five times in the vocabulary (to avoid uncommon misspellings)
   \item a one-hot encoding of categorical labels (i.e., values for the attributes \textit{``contractual choice''}, \textit{``legal form''}, \textit{``municipality''}, \textit{``date''})
   \item the numeric values (\textit{``value''})
\end{itemize}
Then the input matrix, i.e., the matrix that consists of the initial features for each input tender data, is compressed with \textit{Singular Value Decomposition} (SVD) to obtain a dense representation.
SVD was preferred over \textit{Principal Component Analysis} (PCA) due to better performances in processing sparse arrays (such as tf-idf matrices) based on our preliminary results not included in this work.
The result is a dense vector, referred to \textit{SVD} in the rest of this work.
Observe that, differently from our method, the baseline approach to feature extraction requires that the full input is pre-processed to compute the dense representations.
Then, the taxonomy is explored in three strategies:

\paragraph{Big-bang} One classifier is trained to predict a confidence score for all labels, independently from their position in the taxonomy.
As a classifier, we use a random forest\footnote{Using \href{https://scikit-learn.org/stable/}{scikit-learn} implementation; the parameter \texttt{class\_weight} is set to \texttt{balanced} to mitigate the imbalance problem presented above.}.
We refer to this experiment as \textit{RF with SVD (big-bang)}.
This is the only model tested here to avoid taxonomy exploration since it is predicted all at once; therefore we can consider this baseline also as a ``flat'' classifier.
 
\paragraph{Top-down} A classifier is trained for each non-leaf node to predict a confidence score for its descendants.
The exploration of the taxonomy is from the root to leaves and stops when either a leaf is reached or stop conditions are met.  
This approach also uses random forest models as classifiers and therefore we refer to it as \textit{RF with SVD (top-down)}.
 
\paragraph{Per node} This approach uses a different classifier for every node in the taxonomy to predict whether the input belongs to that very node or not.
Exploration of the taxonomy proceeds as in the \textit{per-node} approach, but when a node is explored, its node-specific Boolean classifier is invoked.
Observe that this approach is the most similar to ours, but our approach uses a single model capable of computing a score also for unseen classes.
Due to computational reasons, we preferred to use a logistic regression for this approach.
So, we refer to this experiment as \textit{LR with SVD (per-node)}.

The differences between the baselines are summarized in Table \ref{tab:baseline-diff}. \newline
For all the baselines we adopt the same stopping conditions used in our model: we used a small feed-forward neural network with one hidden layer that takes as input distribution descriptive indexes of the probability scores (maximum and mean values, variance, kurtosis, skewness, difference between the first score and second one, and norm of the output vector) and yields an output in the interval $[0, +1]$.
This approach is an evolution of the one presented by \cite{WANG2022644} that uses logistic regression and is demonstrated to be effective for the task.
Since we are not interested in analyzing this model, but only the classification, we do not report any statistics about this component.
Another sub-task that has been willingly ignored is the extraction of the best candidate from the results: it was just selected in a greedy way choosing the one with the highest score.

\begin{table}[h!]
    \centering
    \begin{tabular}{|l|l|l|}
         \hline
         Approach & Exploration & Label range \\
         \hline
         \textbf{big bang} & no exploration    & all-at-once \\
         \textbf{top-down} & root $\to$ leaves & one per parent (over its descents) \\
         \textbf{per-node} & root $\to$ leaves & one per label (boolean) \\
         \hline
    \end{tabular}
    \caption{Differences between baseline strategies.}
    \label{tab:baseline-diff}
\end{table}

Those approaches were preferred over any LM-based classifiers due to their simplicity: at the end of the process, we have more than $400$ models for the last two experiments, reaching only 15\% of the taxonomy possible sub-trees.

\subsection{HCE Configuration}
Our proposed zero-shot model was tested in two modalities.
In stand-alone mode, i.e., our Hierarchical Cross-encoder (\textit{HCE}) as described in Section~\ref{sec:model} and as an additional prediction layer on top of the \textit{RF with SVD (top-down)} approach (the best baseline according to our results).
This experiments aims to evaulate if a combination of the two may speed-up the zero-shot classification prediction, since HCE is slower compared to the baselines.
Since ranges are different, however, the missing label probabilities are then compared with the most probable label according to the baseline model, and the higher is taken. 

\subsection{Measures}


To evaluate the performance of all models we used a precision measure proposed to evaluate hierarchical classification in~\cite{kiritchenko2005functional}.

\begin{align}
\label{hc_precision}
hP(y, \hat{y}) &= \frac{\sum_i{|y_i \cap \hat{y}_i|}}{\sum_i{|\hat{y}_i|}} \\
\label{hc_recall}
hR(y, \hat{y}) &= \frac{\sum_i{|y_i \cap \hat{y}_i|}}{\sum_i{|y_i|}}
\end{align}

Those formulas calculate the number of common ancestors between ground truth and the predicted class, dividing it either for the predicted label depth ($hP$) or for the ground truth label depth ($hR$).
Although those formulas were originally proposed for micro-precision and micro-recall, we propose to use them as truth scores for the single predictions and aggregate them by ground truth to compute macro scores.
In this way, it becomes possible to test macro-aggregated results by calculating the average truth level per class.
Observe that even a model incapable of  zero-shot classification may reach a positive score since this score is positive if an observation with some ancestors in common with the ground truth is predicted (e.g. a model has never seen a cat, but recognizes it is a mammal).


\subsection{Results}

Table \ref{tab:results} reports hierarchical precision as defined in Equation \ref{hc_precision} for the four experiments.
\begin{table}[h!]
    \centering
    \begin{tabular}{|l|c|c|c|}
        \hline
         Approach & Micro & Macro & $\varrho$ \\
         \hline
         RF with SVD (big-bang) & .2174          & .1190          & \textbf{.2074} \\
         RF with SVD (top-down) & \textbf{.5399} & .3246          & .3109 \\
         LR with SVD (per-node) & .1913          & .0647          & .3234 \\
         \hline
         RF with SVD + HCE (top-down)     & .5334          & .3226          & .3102 \\
         HCE (top-down)                & .5334          & \textbf{.3274} & .2367 \\
         \hline
    \end{tabular}
    \caption{Hierarchical Precision@1 of the experiments. $\varrho$ is the Pearson correlation between label precision and frequency in the training set.}
    \label{tab:results}
\end{table}


We notice that \textit{top-down} baseline model and our HCE achieve comparable performance: the \textit{RF with SVD (top-down)} baseline surpasses HCE in micro-precision by less than $1\%$.
However, the correlation between class macro-precision and class support (column $\varrho$) suggests that performances are dependent on the class frequency in the training data, despite the model mitigation.
Performances of the other two baselines are poor in comparison to our approach: the \textit{RF with SVD (big-bang)} model has over-fit training data (low correlation is due to a random choice of the target variable, mitigated by the model, and not due to a good generalization).
The \textit{LM with SVD (per-node)} approach instead has the worst performance at all.
Our model has a low correlation (the lowest, with the exclusion of the \textit{big-bang} baseline), suggesting better performances with unseen or uncommon labels.
Our model in conjunction with the baseline, instead, reaches lower performances compared to both its sub-components, suggesting poor performances.

To investigate the zero-shot capability of our model, we consider 22 classes that are included in the test set but not in the train set, and therefore were never seen by any model: results are reported in Table \ref{tab:zero-shot}.
On the unseen classes, the zero-shot model (\textit{HCE}) performs quite similarly compared to the rest of the dataset.
A t-test over the two distributions does not exclude the hypothesis that the two results are identical in mean and therefore we can state that our model has similar performances over previously seen and unseen labels.
As a consequence, it is a valid zero-shot approach that relies only on (short) label descriptions.

\begin{table}[h!]
    \centering
    \begin{tabular}{|l|c|c|l|}
        \hline
         Approach & Seen & Unseen & p-value \\
         \hline
         RF with SVD (big-bang) & .1203         & .0474              & .0356  \\
         RF with SVD (top-down) & \textbf{.3304}          & .0192              & $<$ 1e-4 \\
         LR with SVD (per-node) & .0647         & .0000              & .0227    \\
         \hline
         RF with SVD + HCE (top-down)     & .3284          & .0192              & $<$ 1e-4  \\
         HCE (top-down)                & .3273 & \textbf{.3333} & \textbf{.9593}        \\
         \hline
    \end{tabular}
    \caption{Hierarchical macro Precision@1 divided in previously seen and unseen labels. The only two zero-shot models are the HCE ones: our approach can achieve good performances in the zero-shot task and we can not exclude the hypothesis that it has similar performances on both seen and unseen labels (the third column shows p-values for $H_0: seen = unseen$).}
    \label{tab:zero-shot}
\end{table}

The conjunction of the two approaches (\textit{RF with SVD + HCE}) instead obtains the same low performances as the baseline it is composed of.
This result and those reported in Table \ref{tab:results} show that the zero-shot model can not be easily used as is in conjunction with an existing model.
Integrating the two models may require additional effort which is beyond the scope of the present work.
\newline

Since our model yields a list of candidates sorted by score, we are interested also in analyzing how results improve considering more labels than the first one and how results are distributed through the taxonomy.

\begin{table}[h!]
    \centering
    \begin{tabular}{|l|c|cc|c|}
         \hline
           & Micro & seen & unseen & $\varrho$ \\
         \hline
         1 & .5334 & .3273 & .3333 & .2367 \\
         2 & .5893 & .3944 & .4333 & .2122 \\
         3 & .5978 & .4074 & .4333 & .2053 \\
         4 & .6005 & .4107 & .4333 & .2031 \\
         5 & .6017 & .4116 & .4333 & .2028 \\
         \hline
    \end{tabular}
    \caption{Considering other retrieved candidates, the model precision has a significant improvement (initially also for unseen labels). A good choice may be considering the first two or three results; considering the following candidates retrieved by the algorithm does not improve the final result for new labels.}
    \label{tab:precision_k_1_5}
\end{table}

Table \ref{tab:precision_k_1_5} reports results calculated considering the best-suited output for the classification between the best $k$ retrieved candidates: there is considerable growth considering the first three candidates (especially for unseen classes).
Also, the coefficient $\varrho$ decreases, showing that if no unseen classes are found, at least under-represented ones are better classified.

\begin{table}[h]
    \centering
    \begin{tabular}{|r|cc|cc|}
         \hline
               & \multicolumn{2}{c|}{micro} & \multicolumn{2}{c|}{macro} \\
         depth & seen & unseen & seen & unseen \\
         \hline
         root & .5203 & .4688 & .5172 & .4598 \\
         1    & .5964 & .5197 & .3784 & .3817 \\
         2    & .5992 & .1828 & .3315 & .0920 \\
         3    & .5506 & .3333 & .3175 & .5000 \\
         4    & .5342 & .3333 & .3282 & .4000 \\
         5    & .5335 & .2143 & .3302 & .2500 \\
         6    & .5334 & .3000 & .3273 & .3333 \\
         \hline         
    \end{tabular}
    \caption{Results per label depth: results deteriorate with the growing of granularity; unseen classes deteriorate faster than seen classes.}
    \label{tab:per_depth}
\end{table}

Finally, we analyze how the performances for seen and unseen classes are distributed through the taxonomy: results are reported in Table \ref{tab:per_depth}.
We can see that, for previously seen classes results are quite stable, while, on the other hand, new classes are more variable.
However, the lowest p-value for the hypothesis that the two distributions are equal in means (for $k = 2$, $pval = 0.1822$) can never be rejected for any reasonable value.
From those results, we can notice a significant drop in macro-precision at depth $1$, meaning that root labels are easier to recognize compared to more specific labels.
Despite the low (yet positive) correlation between label support and performance, future works may consider different sampling strategies for the training procedure, prioritizing deeper ones.

\section{Conclusions}
In this work we proposed a BERT-based zero-shot method to make document classification over a taxonomy whose labels are described in natural language; we tested it on the \textit{Common Procurement Vocabulary} taxonomy, using industry data in the Italian language.
Our method, consisting in a \textit{Hierarchical Cross-Encoder}, works for single-label classification without the need for output latent representations: it is not always possible nor trivial to obtain a valid representation of unseen labels.
We tested our proposal against three baselines composed of the three main approaches in literature; we were not able to test it against other zero-shot models due to some peculiarity of the taxonomy, which has short label descriptions (even one word long) and negations.

Results show that our model has good generalization capabilities also for previously unseen classes, with very similar performances.
Considering the three more likely candidates, results drastically improve especially for new labels.
However, despite the promising results, our model is quite slow compared to the baseline: we tested if a simple integration of our model with an existing pipeline (here represented by the best-performing baseline) may be successful but with poor results.
Future works may explore strategies in how to speed up the zero-shot model to achieve comparable inference time.

The work presented in this paper has been partially funded by enRichMyData (HE 101070284) and by the PRIN 2022 Project ``Discount quality for responsible data science: Human-in-the-Loop for quality data''.

\printbibliography

\end{document}